# A new closed-loop output error method for parameter identification of robot dynamics

M. Gautier [(1)], A. Janot [(2)] and P-O Vandanjon [(3)]

[(1)] Université de Nantes,
IRCCyN, 1, rue de la Noë - BP 92 101 - 44321 Nantes Cedex 03, France
[(2)] HAPTION S.A,
Atelier Relais de Soulgé Route de Laval, 53210 Soulgé sur Ouette, France
[(3)] Laboratoire Central des Ponts et Chaussées,
Route de Bouaye BP 4129, 44341Bouguenais Cedex, France

*Abstract*—Off-line robot dynamic identification methods are mostly based on the use of the inverse dynamic model, which is linear with respect to the dynamic parameters. This model is sampled while the robot is tracking reference trajectories that excite the system dynamics. This allows using linear least-squares techniques to estimate the parameters. The efficiency of this method has been proved through the experimental identification of many prototypes and industrial robots. However, this method requires the joint force/torque and position measurements and the estimate of the joint velocity and acceleration, through the bandpass filtering of the joint position at high sampling rates. The proposed new method requires only the joint force/torque measurement. It is a closed-loop output error method where the usual joint position output is replaced by the joint force/torque. It is based on a closed-loop simulation of the robot using the direct dynamic model, the same structure of the control law, and the same reference trajectory for both the actual and the simulated robot. The optimal parameters minimize the 2-norm of the error between the actual force/torque and the simulated force/torque. This is a non-linear least-squares problem which is dramatically simplified using the inverse dynamic model to obtain an analytical expression of the simulated force/torque, linear in the parameters. A validation experiment on a 2 degree-of-freedom direct drive robot shows that the new method is efficient.

## I. Introduction

THE usual identification method based on the inverse dynamic identification model (IDIM) and least-squares (LS) technique has been successfully applied to identify inertial and friction parameters of several robotic prototypes and industrial robots [1], [2], [3], [4], [5], [6], [7], [8], [9], [10], [11], [12], [13], [14], [15], amongst others. Good results can be obtained provided a well-tuned derivative bandpass filtering of joint position to calculate the joint velocities and accelerations is used.

Another approach is to minimize a quadratic error between an actual output and a simulated output of the system, assuming both the actual and simulated systems have the same input. This is known as an output error (OE) identification method [16], [17]. The optimal values of the parameters are calculated using non-linear programming algorithms to solve a non-linear least-squares problem. The output is given by a state-space model output equation, which is typically the joint position for mechanical systems. Difficulties arise from the choice of initial conditions, resulting in multiple, local solutions [18]. The OE method has been used to identify electrical parameters of a synchronous machine, and a comparison with the IDIM-LS method showed very similar results [19].

Both IDIM and OE methods require the joint position and the joint force/torque measurements.

The proposed new identification method needs only the joint force/torque measurements. It is based on a closed-loop simulation using the direct dynamic model while the optimal parameters minimize the 2-norm of the error between the actual force/torque and the simulated force/torque, assuming the same control law and the same reference trajectory. This non-linear least-squares problem is dramatically simplified using the inverse dynamic model to formulate the simulated force/torque as an algebraic function linear in relation to the parameters. This paper describes the new identification method and experimental results obtained using a 2 DOF robot.

A condensed version of this work has been presented in [20]. This paper contains detailed proofs to enlighten the theoretical understanding of the method and gives additional experimental results to show the practical efficiency of the method.

The paper is organized as follows: section II reviews the usual identification technique of the dynamic parameters of the robot. Section III presents the output error method. The new identification method is presented in section IV. The modeling of the SCARA prototype robot is presented in section V. This direct drive prototype is very well suitable for the study of the method because it emphasizes non linear coupling while it is divided by the squared high gear ratio for industrial robots. The experimental results are given in section VI. Finally, section VII is the conclusion.

## II. IDIM: INVERSE DYNAMIC IDENTIFICATION MODEL TECHNIQUE

The inverse dynamic model (IDM) of a rigid robot composed of $n$ moving links calculates the motor torque vector $\tau_{idm}$, as a function of the generalized coordinates and their derivatives. It can be obtained from the Newton-Euler or the Lagrangian equations [13], [21]. It is given by the following relation:

$$\tau_{idm} = M(q)\ddot{q} + N(q,\dot{q}) \qquad (1)$$

Where $q$, $\dot{q}$ and $\ddot{q}$ are respectively the $(n\mathrm{x}1)$ vectors of generalized joint positions, velocities and accelerations, $M(q)$ is the $(n\mathrm{x}n)$ robot inertia matrix, and $N(q,\dot{q})$ is the $(n\mathrm{x}1)$ vector of centrifugal, Coriolis, gravitational and friction forces/torques.

The choice of the modified Denavit and Hartenberg frames attached to each link allows a dynamic model that is linear in relation to a set of standard dynamic parameters, $\chi_{st}$ [3], [22]:

$$\tau_{idm} = IDM_{st}(q,\dot{q},\ddot{q})\chi_{st} \qquad (2)$$

Where $IDM_{st}(q,\dot{q},\ddot{q})$ is the $(n\mathrm{x}N_s)$ jacobian matrix of $\tau_{idm}$, with respect to the $(N_s\mathrm{x}1)$ vector $\chi_{st}$ of the standard parameters given by:

$$\chi_{st} = \begin{bmatrix} \chi_{st}^{1\,\mathrm{T}} & \chi_{st}^{2\,\mathrm{T}} & \cdots & \chi_{st}^{n\,\mathrm{T}} \end{bmatrix}^{\mathrm{T}}$$

with:

$$\chi_{st}^{j} = \begin{bmatrix} XX_j & XY_j & XZ_j & YY_j & YZ_j & ZZ_j & MX_j & MY_j & MZ_j & M_j & Ia_j & Fv_j & Fc_j & \tau_{off_j} \end{bmatrix}^{\mathrm{T}} \quad (3)$$

where:
- $XX_j$, $XY_j$, $XZ_j$, $YY_j$, $YZ_j$, $ZZ_j$ are the six components of the inertia matrix, $^jJ_j$, of link j at the origin of frame j,
- $MX_j$, $MY_j$, $MZ_j$ are the components of the first moments, $^jMS_j$, of link $j$,
- $M_j$ is the mass of link $j$,
- $Ia_j$ is a total inertia moment for rotor and gears of actuator $j$.
- $Fv_j$, $Fc_j$ are the viscous and Coulomb friction parameters of joint j.
- $\tau_{off_j} = Of_{Fsj} + Of_{tj}$ is an offset parameter where $Of_{Fsj}$ is the dissymmetry of the Coulomb friction with respect to the sign of the velocity and $Of_{tj}$ is due to the current amplifier offset which supplies the motor.
- $N_s = 14*n$ is the number of standard parameters.

The base parameters are the minimum number of dynamic parameters from which the dynamic model can be calculated. They are obtained from the standard inertial parameters by eliminating those which have no effect on the dynamic model, and by regrouping some others by means of linear relations. They can be determined using simple closed-form rules [22] or a numerical method based on the QR decomposition [23].

The minimal inverse dynamic model can be written as:

$$\tau_{idm} = IDM(q, \dot{q}, \ddot{q}) \chi \quad (4)$$

Where:

$IDM(q, \dot{q}, \ddot{q})$ is the $(n\mathrm{x}b)$ matrix of the minimal set of basis functions of the rigid body dynamics, (5)

$\chi$ is the $(b\mathrm{x}1)$ vector of the $b$ base parameters. (6)

Because of perturbations due to noise measurement and modeling errors, the actual force/torque $\tau$ differs from $\tau_{idm}$ by an error, $e$, such that:

$$\tau = \tau_{idm} + e = IDM(q, \dot{q}, \ddot{q}) \chi + e \quad (7)$$

Equation (7) represents the Inverse Dynamic Identification Model (IDIM).

We consider the off-line identification of the base dynamic parameters $\chi$, given measured or estimated off-line data for $\tau$ and $(q, \dot{q}, \ddot{q})$, collected while the robot is tracking some planned trajectories.

Usually, the signals available from the robot controller are the joint position measurement and the $(n\text{x}1)$ control signal vector $v_\tau$, calculated according to the control law.

Then $(q, \dot{q}, \ddot{q})$ in (7) are estimated with $(\hat{q}, \hat{\dot{q}}, \hat{\ddot{q}})$, respectively, obtained by bandpass filtering the measure of $q$ [9]. The derivatives are calculated off-line without phase shift, using a central difference algorithm of the lowpass filtered position $\hat{q}$. The filtered position $\hat{q}$ is calculated off-line with a non causal zero-phase digital filter by processing the input data, $q$, through a lowpass Butterworth filter in both the forward and reverse direction using the *filtfilt* procedure from Matlab.

The control signal, $v_\tau$, is connected to the input current reference of the current closed-loop of the amplifiers which supplies the motors. Assuming that the current closed-loop has a bandwidth greater than 500Hz, then its transfer function is equal to its static gain, $K_c$, in the frequency range (less than 10Hz) of the rigid robot dynamics. Then, the actual force/torque, $\tau$, is calculated with the relation:

$$\tau = g_\tau v_\tau \tag{8}$$

where:

$g_\tau$, is the $(n\text{x}n)$ diagonal matrix of the drive gains,

with:

$$g_\tau = K_r \, K_c \, K_\tau \tag{9}$$

where:

$K_r$, is the $(n\text{x}n)$ gear ratios diagonal matrix of the joint drive chains ($\dot{q}_m = K_r \dot{q}$, with $\dot{q}_m$, the velocity on the motor side),

$K_c$, is the $(n\text{x}n)$ static gains diagonal matrix of the current amplifiers,

$K_\tau$, is the $(n\text{x}n)$ diagonal matrix of the electromagnetic motor torque constants.

Those parameters have a priori values, given by manufacturers, which can be checked with special tests [24].

The inverse dynamic identification model (IDIM) (7) is calculated at a frequency measurement $f_m$, using samples of $(\hat{q}, \hat{\dot{q}}, \hat{\ddot{q}})$ to calculate $IDM(\hat{q}, \hat{\dot{q}}, \hat{\ddot{q}})$ and samples of $v_\tau$ to calculate $\tau$ with (8), at different times $t_k$, $k = 1,...,n_m$, while the robot is tracking a reference trajectory $(q_r, \dot{q}_r, \ddot{q}_r)$, during the time length $T_{obs}$, of the trajectory.

The equations of each joint are regrouped together on all the trajectory to get an over-determined linear system such that:

$$Y_{fm}(\tau) = W_{fm}(\hat{q},\hat{\dot{q}},\hat{\ddot{q}})\chi + \rho_{fm} \qquad (10)$$

With:

$$Y_{fm}(\tau) = \begin{bmatrix} Y_{fm}^1 \\ ... \\ Y_{fm}^n \end{bmatrix}, \; Y_{fm}^j = \begin{bmatrix} \tau_j(t_1) \\ ... \\ \tau_j(t_{n_m}) \end{bmatrix} \qquad (11)$$

$$W_{fm}(\hat{q},\hat{\dot{q}},\hat{\ddot{q}}) = \begin{bmatrix} W_{fm}^1 \\ ... \\ W_{fm}^n \end{bmatrix}, \; W_{fm}^j = \begin{bmatrix} IDM^j(\hat{q}(t_1),\hat{\dot{q}}(t_1),\hat{\ddot{q}}(t_1)) \\ ... \\ IDM^j(\hat{q}(t_{n_m}),\hat{\dot{q}}(t_{n_m}),\hat{\ddot{q}}(t_{n_m})) \end{bmatrix} \qquad (12)$$

where:

$IDM^j(\hat{q}(t_k),\hat{\dot{q}}(t_k),\hat{\ddot{q}}(t_k))$ is the jth row of the $(n \times b)$ matrix of the basis functions, $IDM(\hat{q}(t_k),\hat{\dot{q}}(t_k),\hat{\ddot{q}}(t_k))$, (5),

$Y_{fm}^j$ and $W_{fm}^j$ represent the $n_m$ equations of joint $j$,

$n_m = T_{obs} * f_m$ is the number of sample measurements.

The notation $W_{fm}(IDM(\hat{q},\hat{\dot{q}},\hat{\ddot{q}})) = W_{fm}(\hat{q},\hat{\dot{q}},\hat{\ddot{q}})$, will be used to recall that $W_{fm}$, is calculated with a sampling of $IDM(\hat{q},\hat{\dot{q}},\hat{\ddot{q}})$.

In order to eliminate high frequency force/torque ripple in $\tau$, and to window the identification frequency range into the model dynamics, a parallel decimation procedure lowpass filters in parallel $Y_{fm}$ and each column of $W_{fm}$ and resamples them at a lower rate, keeping one sample over $n_d$. This parallel decimation can be carried out with the Matlab *decimate* function, where the lowpass filter cut-off frequency is equal to $0.8*f_m/(2*n_d)$.

After the data acquisition procedure and the parallel decimation of (10), we obtain the over-determined linear system:

$$Y(\tau) = W(\hat{q},\hat{\dot{q}},\hat{\ddot{q}})\chi + \rho \qquad (13)$$

where:

- $Y(\tau)$ is the $(r \times 1)$ vector of measurements, built from the actual force/torque $\tau$,
- $W(\hat{q},\hat{\dot{q}},\hat{\ddot{q}})$ is the $(r \times b)$ observation matrix, built from the estimated values $(\hat{q},\hat{\dot{q}},\hat{\ddot{q}})$ of $(q, \dot{q}, \ddot{q})$.

- $\rho$ is the $(r \times 1)$ vector of errors.
- $r = n * n_m / n_d$ is the number of rows in (13).

In $Y$ and $W$, the equations of each joint are grouped together such that:

$$Y = \begin{bmatrix} Y^1 \\ ... \\ Y^n \end{bmatrix}, \quad W = \begin{bmatrix} W^1 \\ ... \\ W^n \end{bmatrix} \tag{14}$$

where $Y^j$ and $W^j$ represent the $n_m / n_d$ equations of joint $j$.

The ordinary LS (OLS) solution $\hat{\chi}$ minimizes the squared 2-norm $\|\rho\|^2$ of the vector of errors.

Using the base parameters and tracking "exciting" reference trajectories [25], we get a full rank and well conditioned matrix $W$. The LS solution $\hat{\chi}$ is given by:

$$\hat{\chi} = \left( (W^T W)^{-1} W^T \right) Y = W^+ Y \tag{15}$$

It is computed using the QR factorization of $W$. Standard deviations $\sigma_{\hat{\chi}_i}$, are estimated using classical results from statistics under the assumptions that $W$ is a deterministic matrix, according to the data filtering procedure described above, and $\rho$, is a zero-mean additive independent Gaussian noise, with a covariance matrix $C_{\rho\rho}$, such that:

$$C_{\rho\rho} = E(\rho \rho^T) = \sigma_\rho^2 I_r \tag{16}$$

where $E$ is the expectation operator and $I_r$, the $(r \times r)$ identity matrix.

An unbiased estimation of the standard deviation $\sigma_\rho$ is:

$$\hat{\sigma}_\rho^2 = \|Y - W\hat{\chi}\|^2 / (r - b) \tag{17}$$

The covariance matrix of the estimation error is given by:

$$C_{\hat{\chi}\hat{\chi}} = E[(\chi - \hat{\chi})(\chi - \hat{\chi})^T] = \hat{\sigma}_\rho^2 (W^T W)^{-1} \tag{18}$$

$\sigma_{\hat{\chi}_i}^2 = C_{\hat{\chi}\hat{\chi}}(i,i)$ is the i$^{th}$ diagonal coefficient of $C_{\hat{\chi}\hat{\chi}}$. The relative standard deviation $\%\sigma_{\hat{\chi}_{ri}}$ is given by:

$$\%\sigma_{\hat{\chi}_{ri}} = 100 \sigma_{\hat{\chi}_i} / |\hat{\chi}_i|, \text{ for } |\hat{\chi}_i| \neq 0 \tag{19}$$

The OLS can be improved by taking into account different standard deviations on joint $j$ equations errors [9]. Each equation of joint $j$ in (13), (14), is weighted with the inverse of the standard deviation of the error calculated from OLS solution of the equations of joint $j$, given by:

$$Y^j(\tau_j) = W^j \left( IDM^j(\hat{q}, \hat{\dot{q}}, \hat{\ddot{q}}) \right) \chi + \rho^j \tag{20}$$

This weighting operation normalises the errors in (13) and gives the weighted LS (WLS) estimation of the parameters.

This identification method is illustrated in Fig. 1.

Compared with the OE method described in the following section III, the use of IDIM, which is an analytical function of $(q,\dot{q},\ddot{q})$, is particularly interesting because it does not require the integration of the direct dynamic model (21). Moreover, $\hat{\chi}$ is a one step linear LS solution which does not need initial conditions. However, the calculation of the velocities and accelerations are required using well-tuned bandpass filtering of the joint position [9].

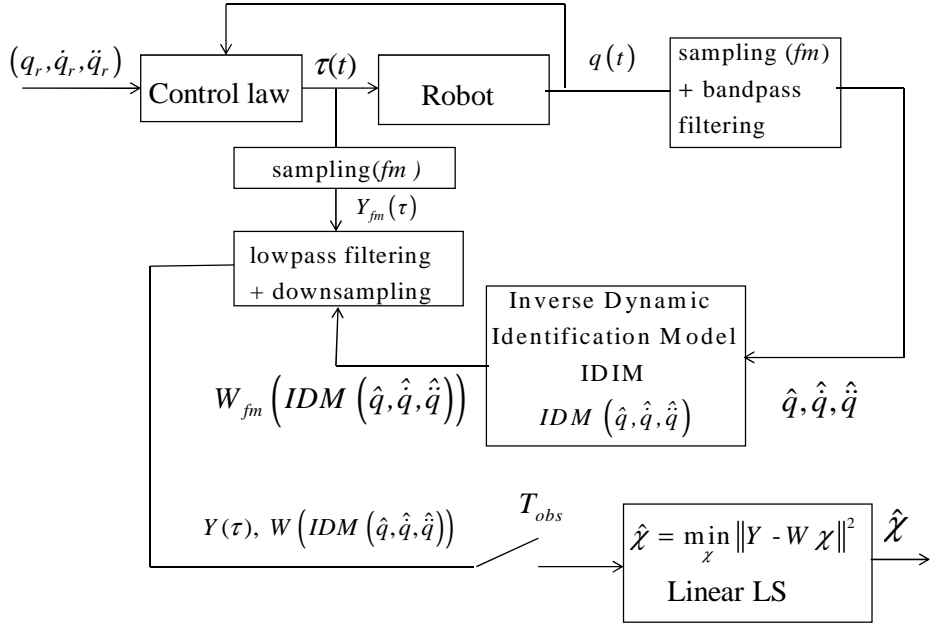

Fig. 1. IDIM LS identification scheme.

### III. THE OUTPUT ERROR METHOD (OE)

The OE identification methods minimize a quadratic error between an actual output $y$, and a simulated output $y_s$, of the system, assuming both the actual and the simulated systems have the same input. This approach can be implemented in an open-loop form, [17], [26], or in a closed-loop form, [27], [28]. Considering a closed-loop controlled robot, the input, in the open loop scheme shown in Fig. 2, is the actual force/torque $\tau$, and the input, in the closed-loop scheme shown in Fig. 3, is the reference trajectory $(q_r,\dot{q}_r,\ddot{q}_r)$. Because the open loop simulation of unstable robotic systems is very sensitive to the initial state conditions and to the errors in numerical algorithms which solve the differential equations, it is more suitable to choose the closed-loop form.

In both cases, the output is given by a state-space model output equation. Considering a robot and

This work has been submitted to the IEEE for possible publication. Copyright may be transferred without notice, after which this version may no longer be accessible      p 7 taking the measured joint position as the output, the actual output is, $y = q$, and the simulated output is, $y_s = q_{ddm}$, as shown in Fig. 2 and Fig. 3, where $q_{ddm}(t)$, is the solution of the differential equation given by the Direct Dynamic Model (DDM).

The DDM can be obtained by writing the IDM equation (1), as following:

$$M(q_{ddm},\chi) \ddot{q}_{ddm} = \tau_{ddm} - N(q_{ddm}, \dot{q}_{ddm}, \chi) \tag{21}$$

where:

$M(q_{ddm},\chi)$ and $N(q_{ddm}, \dot{q}_{ddm}, \chi)$ depend on an estimation of the base parameters $\chi$,

$\tau_{ddm}$, is the force/torque input of the DDM.

The function $q_{ddm}(t,\chi)$, is the result of the integration of the linear implicit differential equation (21) which can be written as a non-linear state-space model:

$$G(x_s)\dot{x}_s = f(x_s, u_s) \tag{22}$$

where:

$x_s = \begin{bmatrix} q_{ddm} \\ \dot{q}_{ddm} \end{bmatrix}$, is the $(2*n \times 1)$ state-space vector,

$u_s = \tau_{ddm}$, is the $(n \times 1)$ control input,

$$G(x_s) = \begin{bmatrix} I_n & 0_{n \times n} \\ 0_{n \times n} & M(q_{ddm},\chi) \end{bmatrix}, f(x_s, u_s) = \begin{bmatrix} \dot{q}_{ddm} \\ \tau_{ddm} - N(q_{ddm}, \dot{q}_{ddm}, \chi) \end{bmatrix} \tag{23}$$

where, $0_{n \times n}$, is a $(n \times n)$, matrix of zeros.

The linear output equation is given by:

$$y_s = C_s x_s + D_s u_s \tag{24}$$

Taking the measure of joint position as the output, $y_s = q_{ddm}$, we get:

$C_s = \begin{bmatrix} I_n & 0_{n \times 2*n} \end{bmatrix}$, is the, $(n \times 2*n)$, output matrix, \hfill (25)

$D_s = 0_{n \times n}$, is the, $(n \times n)$, direct feedthrough matrix. \hfill (26)

Hence, for robotic systems, an OE identification method is based on the integration of the Direct Dynamic Model.

The optimal solution $\hat{\chi}$, minimizes the quadratic criterion $J(\chi)$, given by:

$$J(\chi) = \|Y_s - Y\|^2 = (Y_s - Y)^T (Y_s - Y) \tag{27}$$

where:

$Y$ and $Y_s$, are vectors obtained by filtering the vectors of samples $Y_{fm}$ and $Y_{Sfm}$, respectively, where the equations of each joint are grouped together, with:

$$Y_{fm} = \begin{bmatrix} Y_{fm}^1 \\ ... \\ Y_{fm}^n \end{bmatrix},\ Y_{fm}^j = \begin{bmatrix} q_j(t_1) \\ ... \\ = q_j(t_{n_m}) \end{bmatrix},\ Y_{Sfm} = \begin{bmatrix} Y_{Sfm}^1 \\ ... \\ Y_{Sfm}^n \end{bmatrix},\ Y_{Sfm}^j = \begin{bmatrix} q_{ddm}(t_1) \\ ... \\ q_{ddm}(t_k) \end{bmatrix} \tag{28}$$

The minimization of $J(\chi)$, (27), is a non-linear least-squares problem. The estimation of the parameters can be computed using algorithms such as the gradient method, the Newton methods or the Levenberg Marquardt method. These methods are based on a first or second order Taylor's expansion of $J(\chi)$.

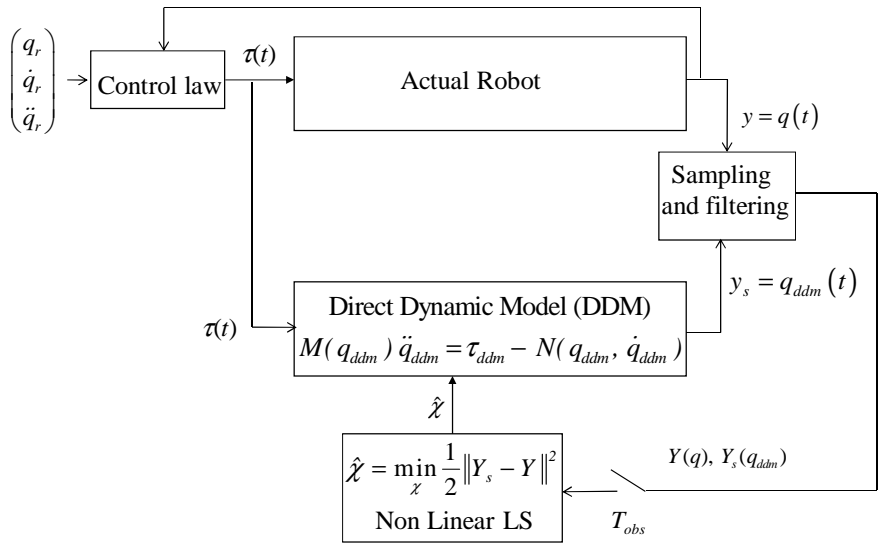

Fig. 2. Open-loop OE identification scheme.

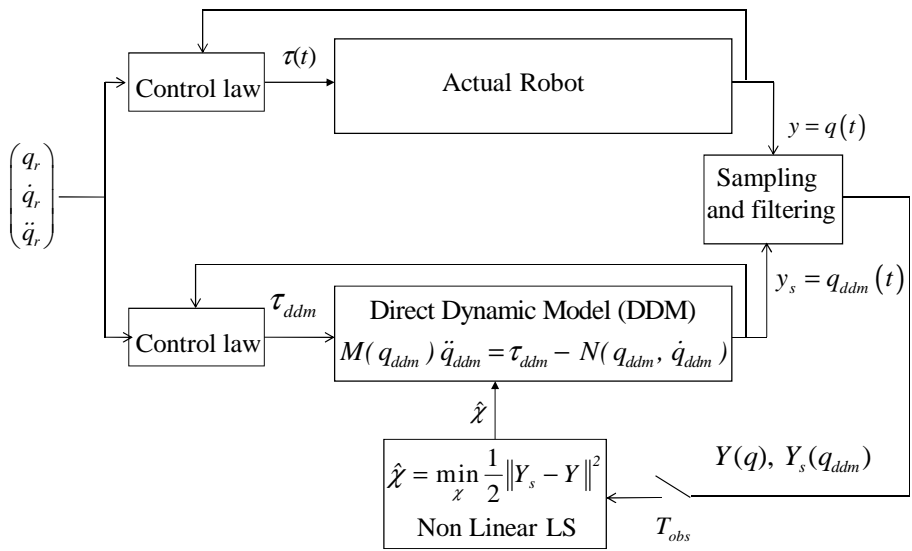

Fig. 3. Closed-loop OE identification scheme.

In [20], we used the Gauss-Newton method to calculate the optimal solution. It is a Newton method where approximations of the gradient and the hessian of $J(\chi)$ are calculated with the jacobian matrix of $y_s$ with respect to $\chi$. The Gauss-Newton regression is a simpler way to calculate the optimal solution [29]. It is based on a Taylor series expansion of $y_s$, at a current estimate $\hat{\chi}^k$, of the parameters at iteration $k$:

$$y_S(\chi^{k+1}) = y_S(\hat{\chi}^k) + \left(\frac{\partial(y_S(\chi))}{\partial \chi}\right)_{\hat{\chi}^k} (\chi^{k+1} - \hat{\chi}^k) + o \tag{29}$$

where:

$$\left(\frac{\partial(y_S(\chi))}{\partial \chi}\right)_{\hat{\chi}^k} = \delta_{y_s/\chi} \tag{30}$$

$\delta_{y_s/\chi}$ is the $(n \mathrm{x} b)$, jacobian matrix of $y_s$, with respect to $\chi$, evaluated at $\hat{\chi}^k$.

Each coefficient of $\delta_{y_s/\chi}$, defines a sensitivity function.

These sensitivity functions characterize the variation of the output function $y_S$, with respect to a variation of the parameter $\chi$. The sensitivity functions are the solutions of a differential system calculated from (21). However, this technique is more time-consuming compared to the IDIM method. Indeed, the DDM and the sensitivity functions must be integrated many times at each step of the iterative non-linear optimization method. Moreover, it is necessary to have good initial conditions in order to avoid multiple and local solutions.

Let us define:

$$y = y_S(\chi^{k+1}) + e \tag{31}$$

From (29), it becomes:

$$y - y_s(\hat{\chi}^k) = \delta_{y_s/\chi}(\chi^{k+1} - \hat{\chi}^k) + (o + e) \tag{32}$$

An over-determined linear system is obtained by filtering and sampling (32) over the time window $T_{obs}$:

$$\Delta Y = W_\delta \Delta \chi^{k+1} + \rho \tag{33}$$

with:

$$\Delta \chi^{k+1} = (\chi^{k+1} - \hat{\chi}^k)$$

$\Delta Y$, $W_\delta$, and $\rho$ are, respectively, the sampling and filtering of $\left(y - y_s(\hat{\chi}^k)\right)$, $\delta_{y_s/\chi}$, and of $(o+e)$.

$\Delta \hat{\chi}^{k+1}$ is the LS solution of (33). This process is iterated with a new estimate, $\hat{\chi}^{k+1} = \hat{\chi}^k + \Delta \hat{\chi}^{k+1}$, until:

$$\frac{\|\rho_{k+1}\| - \|\rho_k\|}{\|\rho_k\|} \leq tol_1, \text{ and, } \max_{i=1,\ldots,b} \left| \frac{\hat{\chi}_i^{k+1} - \hat{\chi}_i^k}{\hat{\chi}_i^k} \right| \leq tol_2 \quad (34)$$

where, $tol_1$ and $tol_2$, are values ideally chosen to be small numbers to get fast convergence with good accuracy.

## IV. DIDIM: DIRECT AND INVERSE DYNAMIC IDENTIFICATION MODEL TECHNIQUE

### A. Theoretical approach

In the OE method as shown in Fig. 3, the actual output is the measured joint position, $y = q$.

We propose to change the output, $y$, from the actual joint position $q$, to the actual joint force/torque $\tau$, and the simulated output $y_s$, from the simulated joint position, $q_{ddm}$, to the simulated joint force/torque, $\tau_{ddm}$. Then, we take $y = \tau$, and $y_s = \tau_{ddm}$, according to Fig. 4.

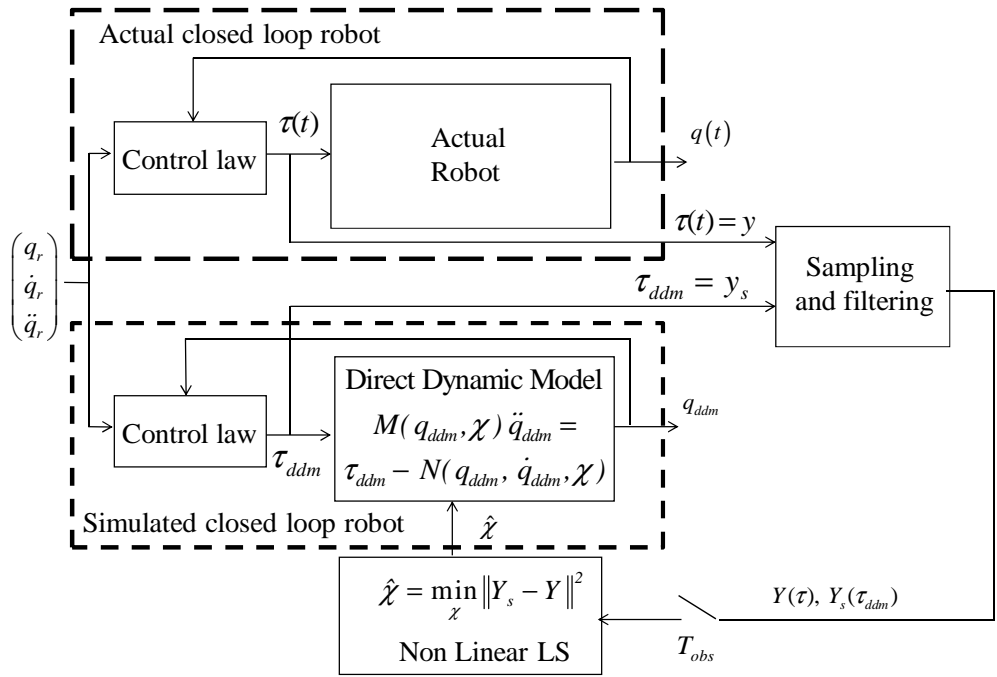

Fig. 4. DIDIM identification scheme.

This means that the output equation (24) of the state-space model (22) reduces to a direct feedthrough equation such as, $y_s = u_s = \tau_{ddm}$.

Then we have $C_s = 0_{n \times 2*n}$, and $D_s = I_n$, in the output equation (24).

 

The optimal solution, $\hat{\chi}$, minimizes the quadratic criterion, $J(\chi)$, (27), where, $Y$, and $Y_s$, are vectors obtained by filtering the vectors of samples, $Y_{fm}$ and $Y_{Sfm}$, respectively, where the equations of each joint are grouped, with:

$$Y_{fm}(\tau) = \begin{bmatrix} Y_{fm}^1 \\ \ldots \\ Y_{fm}^n \end{bmatrix}, \quad Y_{fm}^j = \begin{bmatrix} \tau_j(t_1) \\ \ldots \\ \tau_j(t_{n_m}) \end{bmatrix}, \quad Y_{Sfm} = \begin{bmatrix} Y_{Sfm}^1 \\ \ldots \\ Y_{Sfm}^n \end{bmatrix}, \quad Y_{Sfm}^j = \begin{bmatrix} \tau_{ddm_j}(t_1) \\ \ldots \\ \tau_{ddm_j}(t_{n_m}) \end{bmatrix} \quad (35)$$

This non-linear LS problem is solved by the Gauss-Newton regression as explained in section III.

The input force/torque of the DDM, $\tau_{ddm}$, can be calculated with the analytical expression of the inverse dynamic model (4), such as:

$$y_s(\chi) = \tau_{ddm}(\chi) = \tau_{idm}(\chi) = IDM\left(q_{ddm}(\chi), \dot{q}_{ddm}(\chi), \ddot{q}_{ddm}(\chi)\right)\chi \quad (36)$$

The Taylor series expansion (29), with $y_s = \tau_{ddm}$, at a current estimate, $\hat{\chi}^k$, of the parameters $\chi$, at iteration $k$, is calculated with the jacobian matrix of $\tau_{ddm}(\chi)$, given by:

$$\delta_{y_s/\chi} = \left(\frac{\partial(\tau_{ddm})}{\partial \chi}\right)_{\hat{\chi}^k} = \left(\frac{\partial(\tau_{idm})}{\partial \chi}\right)_{\hat{\chi}^k} = \frac{\partial}{\partial \chi}\left(IDM\left(q_{ddm}(\hat{\chi}^k), \dot{q}_{ddm}(\hat{\chi}^k), \ddot{q}_{ddm}(\hat{\chi}^k)\right)\hat{\chi}^k\right) \quad (37)$$

Then, it becomes:

$$\frac{\partial}{\partial \chi}\left(IDM\left(q_{ddm}(\hat{\chi}^k), \dot{q}_{ddm}(\hat{\chi}^k), \ddot{q}_{ddm}(\hat{\chi}^k)\right)\hat{\chi}^k\right) = IDM\left(q_{ddm}(\hat{\chi}^k), \dot{q}_{ddm}(\hat{\chi}^k), \ddot{q}_{ddm}(\hat{\chi}^k)\right) + \ldots$$
$$\frac{\partial}{\partial \chi}\left(IDM\left(q_{ddm}(\hat{\chi}^k), \dot{q}_{ddm}(\hat{\chi}^k), \ddot{q}_{ddm}(\hat{\chi}^k)\right)\right)\hat{\chi}^k \quad (38)$$

The calculation of the second term on the right side of (38) needs to calculate the expression:

$$\frac{\partial}{\partial \chi}\left(IDM\left(q_{ddm}(\hat{\chi}^k), \dot{q}_{ddm}(\hat{\chi}^k), \ddot{q}_{ddm}(\hat{\chi}^k)\right)\right) = \frac{\partial}{\partial q_{ddm}}\left(IDM\left(q_{ddm}(\hat{\chi}^k), \dot{q}_{ddm}(\hat{\chi}^k), \ddot{q}_{ddm}(\hat{\chi}^k)\right)\right)\frac{\partial q_{ddm}}{\partial \chi} + \ldots$$
$$\frac{\partial}{\partial \dot{q}_{ddm}}\left(IDM\left(q_{ddm}(\hat{\chi}^k), \dot{q}_{ddm}(\hat{\chi}^k), \ddot{q}_{ddm}(\hat{\chi}^k)\right)\right)\frac{\partial \dot{q}_{ddm}}{\partial \chi} + \ldots \quad (39)$$
$$\frac{\partial}{\partial \ddot{q}_{ddm}}\left(IDM\left(q_{ddm}(\hat{\chi}^k), \dot{q}_{ddm}(\hat{\chi}^k), \ddot{q}_{ddm}(\hat{\chi}^k)\right)\right)\frac{\partial \ddot{q}_{ddm}}{\partial \chi}$$

Let us recall that the joint force/torque $y = \tau$, is obtained while the robot is tracking a reference trajectory, $(q_r, \dot{q}_r, \ddot{q}_r)$, with a closed-loop control law. The closed-loop simulation uses the direct dynamic model, the same control law and the same reference trajectory $(q_r, \dot{q}_r, \ddot{q}_r)$, as the actual one, to calculate $y_S$.

In the following section IV.B, we show how to tune the control law of the closed-loop simulation in order to keep the same bandwidth and stability margin as the actual closed-loop for any $\hat{\chi}^k$, obtained at iteration $k$. This assumes for the simulated tracking error to keep close to the actual one for any $\hat{\chi}^k$, that is to say:

$$\left(q_{ddm}(\hat{\chi}^k), \dot{q}_{ddm}(\hat{\chi}^k), \ddot{q}_{ddm}(\hat{\chi}^k)\right) \approx (q, \dot{q}, \ddot{q}) \text{, for any } \hat{\chi}^k \quad (40)$$

This means that $\left(q_{ddm}(\chi), \dot{q}_{ddm}(\chi), \ddot{q}_{ddm}(\chi)\right)$, have little dependence on $\chi$, such that:

$$\frac{\partial q_{ddm}}{\partial \chi} \approx \frac{\partial \dot{q}_{ddm}}{\partial \chi} \approx \frac{\partial \ddot{q}_{ddm}}{\partial \chi} \approx 0$$

Then (39) is simplified as:

$$\frac{\partial}{\partial \chi}\left(IDM\left(q_{ddm}(\hat{\chi}^k), \dot{q}_{ddm}(\hat{\chi}^k), \ddot{q}_{ddm}(\hat{\chi}^k)\right)\right) \approx 0$$

Taking into account this simplification, we have in (38):

$$\frac{\partial}{\partial \chi}\left(IDM\left(q_{ddm}(\hat{\chi}^k), \dot{q}_{ddm}(\hat{\chi}^k), \ddot{q}_{ddm}(\hat{\chi}^k)\right)\right)\hat{\chi}_k \ll IDM\left(q_{ddm}(\hat{\chi}^k), \dot{q}_{ddm}(\hat{\chi}^k), \ddot{q}_{ddm}(\hat{\chi}^k)\right)$$

As a result, the jacobian matrix (37) can be approximated by:

$$\delta_{y_s/\chi} = \frac{\partial}{\partial \chi}\left(IDM\left(q_{ddm}(\hat{\chi}^k), \dot{q}_{ddm}(\hat{\chi}^k), \ddot{q}_{ddm}(\hat{\chi}^k)\right)\hat{\chi}^k\right) \approx IDM\left(q_{ddm}(\hat{\chi}^k), \dot{q}_{ddm}(\hat{\chi}^k), \ddot{q}_{ddm}(\hat{\chi}^k)\right) \quad (41)$$

Each sensitivity function in the jacobian matrix is approximated by an algebraic equation. This is much more simpler than for usual OE method where the sensitivity functions are the solutions of complicated differential equations. This is the reason why it is much more simpler to minimize the error between the measured force/torque and the simulated force/torque than to minimize the error between the actual position and the simulated position.

Taking the approximation (41) of the jacobian matrix into the Taylor series expansion (32), it becomes:

$$y = \tau = y_s(\hat{\chi}^k) + IDM\left(q_{ddm}(\hat{\chi}^k), \dot{q}_{ddm}(\hat{\chi}^k), \ddot{q}_{ddm}(\hat{\chi}^k)\right)\left(\chi^{k+1} - \hat{\chi}^k\right) + (o + e) \quad (42)$$

From (36), it becomes:

$$y_s(\hat{\chi}^k) = \tau_{idm}(\hat{\chi}^k) = IDM\left(q_{ddm}(\hat{\chi}^k), \dot{q}_{ddm}(\hat{\chi}^k), \ddot{q}_{ddm}(\hat{\chi}^k)\right)\hat{\chi}^k \quad (43)$$

Taking (43) in (42), it becomes:

$$y = \tau = IDM\left(q_{ddm}(\hat{\chi}^k), \dot{q}_{ddm}(\hat{\chi}^k), \ddot{q}_{ddm}(\hat{\chi}^k)\right)\chi^{k+1} + (o + e) \quad (44)$$

This is the Inverse Dynamic Identification Model, IDIM, (7), where $(q, \dot{q}, \ddot{q})$ are estimated with $(q_{ddm}, \dot{q}_{ddm}, \ddot{q}_{ddm})$, simulated with $DDM(\hat{\chi}^k)$ (21). At each iteration $k$, the IDIM method is applied as described in section II.

The sampling of (44) at a sampling rate $f_m$, gives an over-determined linear system such as:

$$Y_{fm}(\tau) = W_{\delta fm}(q_{ddm}, \dot{q}_{ddm}, \ddot{q}_{ddm}, \hat{\chi}^k)\chi + \rho_{fm} \qquad (45)$$

With:

$$Y_{fm}(\tau) = \begin{bmatrix} Y_{fm}^1 \\ ... \\ Y_{fm}^n \end{bmatrix}, \quad Y_{fm}^j = \begin{bmatrix} \tau_j(t_1) \\ ... \\ \tau_j(t_{n_m}) \end{bmatrix} \qquad (46)$$

$$W_{\delta fm}(q_{ddm}, \dot{q}_{ddm}, \ddot{q}_{ddm}, \hat{\chi}^k) = \begin{bmatrix} W_{\delta fm}^1 \\ ... \\ W_{\delta fm}^n \end{bmatrix}, \quad W_{\delta fm}^j = \begin{bmatrix} IDM^j(q_{ddm}(t_1), \dot{q}_{ddm}(t_1), \ddot{q}_{ddm}(t_1), \hat{\chi}^k) \\ ... \\ IDM^j(q_{ddm}(t_{n_m}), \dot{q}_{ddm}(t_{n_m}), \ddot{q}_{ddm}(t_{n_m}), \hat{\chi}^k) \end{bmatrix} \qquad (47)$$

The parallel decimation of (45) gives:

$$Y(\tau) = W_{\delta}(q_{ddm}, \dot{q}_{ddm}, \ddot{q}_{ddm}, \hat{\chi}^k)\chi + \rho \qquad (48)$$

The LS solution of (48) gives $\hat{\chi}_{k+1}$, at iteration $k+1$.

This process is iterated until:

$$\frac{\|\rho_{k+1}\| - \|\rho_k\|}{\|\rho_k\|} \leq tol_1, \text{ and, } \max_{i=1,...,b}\left|\frac{\hat{\chi}_i^{k+1} - \hat{\chi}_i^k}{\hat{\chi}_i^k}\right| \leq tol_2,$$

where, $tol_1$ and $tol_2$ are values ideally chosen to be small numbers to get fast convergence with good accuracy.

This new identification method is based on a closed-loop simulation using the direct dynamic model (DDM) while the optimal parameters minimize the 2-norm of the error between the actual force/torque $\tau$, and the simulated force/torque $\tau_{ddm}$, over an observation window time $T_{obs}$. This new technique overcomes the problems of non-linear optimization in OE method, section III, using the IDM to calculate the simulated force/torque vector, $y_s = \tau_{ddm} = \tau_{idm}$. Because this method uses both models DDM and IDM, it is named the DIDIM method: Direct and Inverse Dynamic Identification Models technique.

The DIDIM method with the Gauss-Newton regression is illustrated Fig. 5.

This approach is particularly interesting thanks to the following reasons:

- It needs only the actuator force/torque measurement or estimation,
- It avoids tuning the bandpass filter in the IDIM method by using the integration of the DDM in a closed-loop simulation where the tuning of the bandwidth automatically defines the same frequency

 

range for the dynamics of the actual and of the model to be identified.
- It combines the inverse and the direct dynamic model and validates, in the same identification procedure, both models for computed torque control and for simulation.
- It dramatically simplifies the computation of the matrix of the sensitivity functions which is given by an algebraic equation (the inverse dynamic identification model) whereas it is given by the resolution of a complicated system of differential equations in the usual OE method.

The drawback is that the structure and the tuning of the actual closed-loop control law must be known to be implemented in the closed-loop simulation of the robot. Most often, this is not a real problem, because working on identification for simulation or control of the robot, needs a minimal knowledge on the robot controller.

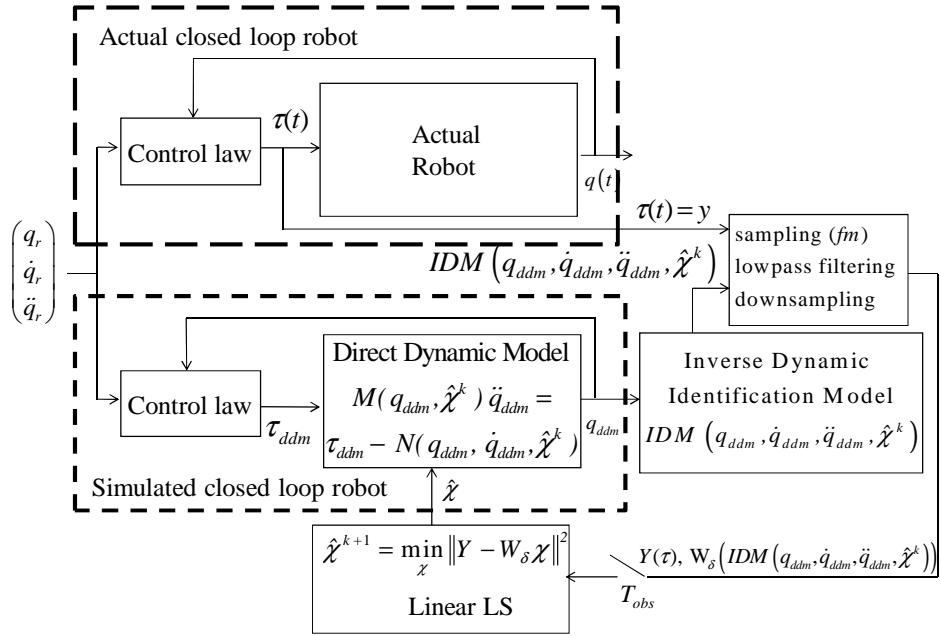

Fig. 5. DIDIM with the Gauss-Newton regression, identification scheme.

B. *Initialization of the algorithm*

A problem is how to choose the initial values $\hat{\chi}^0$.

We can use CAD values, or identified values with the IDIM method, but we show that there is no need at all of a priori values.

We propose an algorithm not sensitive to the initial conditions, which assumes that the condition $(q_{ddm}(\hat{\chi}_k), \dot{q}_{ddm}(\hat{\chi}_k), \ddot{q}_{ddm}(\hat{\chi}_k)) \approx (q, \dot{q}, \ddot{q})$, is satisfied at any iteration $k$, and especially for $k=0$.

This is possible by taking the same control law structure for the actual robot and for the simulated one with the same performances given by the bandwidth, the stability margin or the closed-loop poles.

Because the simulated robot parameters $\hat{\chi}^k$, change at each iteration $k$, the gains of the simulated control law must be updated according to $\hat{\chi}^k$.

For example, let us consider a PD control law for each joint $j$. The inverse dynamic model IDM (1) for the joint $j$, can be written as a decoupled double integrator perturbed by a coupling force/torque, such that:

$$\tau_j = \tau_{idm_j} = \sum_{i=1}^{n} M_{j,i}(q)\,\ddot{q}_i + N_j(q,\dot{q}) = M_{j,j}(q)\,\ddot{q}_j + \sum_{i \neq j}^{n} M_{j,i}(q)\,\ddot{q}_i + N_j(q,\dot{q}) = M_{j,j}(q)\,\ddot{q}_j - p_j \qquad (49)$$

where $p_j$ is considered as a perturbation given by:

$$p_j = -\sum_{i \neq j}^{n} M_{j,i}(q)\,\ddot{q}_i - N_j(q,\dot{q}) \qquad (50)$$

$M_{j,i}(q)$ which depends on $q$, is approximated by a constant inertia moment $J_j$, given by:

$$J_j = ZZ_j + I_{a_j} + \max_{q}\left(M_{j,j}(q) - ZZ_j - I_{a_j}\right) \qquad (51)$$

$J_j$, is the maximum value, with respect to $q$, of the inertia moment around joint $z_j$ axis. This gives the smallest damping value and the smallest stability margin of the closed-loop second order transfer function (55), while $q$ varies.

It can be calculated from a priori CAD values of inertial parameters and must be taken at least as $ZZ_j + I_{a_j}$.

The joint $j$ dynamic model is approximated by a double integrator, where $p_j$, is a perturbation, as following:

$$\ddot{q}_j = \frac{1}{M_{j,j}(q)}\left(\tau_j + p_j\right) \quad \frac{1}{J_j}\left(\tau_j + p_j\right) \qquad (52)$$

Let us consider the joint $j$ PD control of the actual robot which is illustrated Fig. 6:

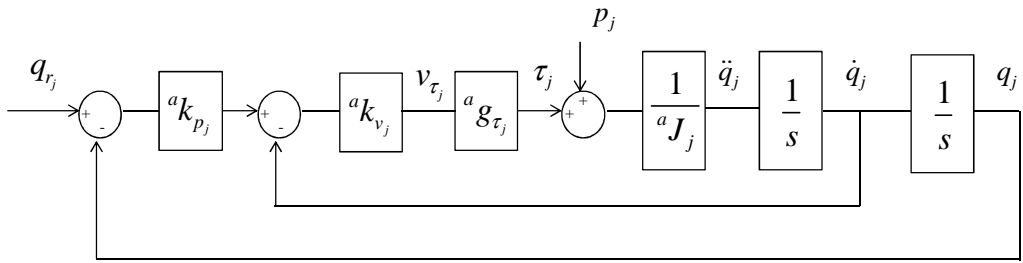

Fig. 6. Joint PD control of the actual robot.

The control input calculated by the robot controller is given by:

$$v_{\tau_j} = {}^a k_{p_j} {}^a k_{v_j} \left( q_{r_j} - q_j \right) - {}^a k_{v_j} \dot{q}_j \tag{53}$$

$v_{\tau_j}$ is the current reference of the current amplifiers which supplies the motor.

The joint $j$, force/torque is given by:

$$\tau_j = {}^a g_{\tau_j} v_{\tau_j} \tag{54}$$

where:

${}^a g_{\tau_j}$ is the actual drive gain, calculated with the actual parameters in (9).

${}^a J_j$ is the actual value of $J_j$.

In order to tune the tracking performances of the reference position $q_{r_j}$, the transfer function $\dfrac{q_{r_j}}{q_j}$ is calculated with $p_j = 0$:

$$H_j = \left( \frac{q_j}{q_{r_j}} \right)_{p_j=0} = \frac{1}{\dfrac{{}^a J_j \, s^2}{{}^a g_{\tau_j} {}^a k_{v_j} {}^a k_{p_j}} + \dfrac{1}{{}^a k_{p_j}} s + 1} = \frac{1}{\dfrac{s^2}{{}^a \omega_{nj}^2} + \dfrac{2 \, {}^a \zeta_j}{{}^a \omega_{nj}} s + 1} \tag{55}$$

where:

${}^a \omega_{nj}$ is the actual natural frequency which characterizes the closed-loop bandwidth,

${}^a \zeta_j$ is the actual damping coefficient which characterizes the closed-loop stability margin, with:

$$ {}^a \omega_{nj} = \sqrt{{}^a k_{p_i} {}^a k_{v_i} \frac{{}^a g_{\tau_j}}{{}^a J_j}} , \qquad {}^a \zeta_j = \frac{1}{2} \sqrt{\frac{{}^a k_{v_i} \, {}^a g_{\tau_j}}{{}^a k_{p_i} \, {}^a J_j}} \tag{56}$$

Then it becomes:

$$ {}^a k_{p_j} = \frac{{}^a \omega_{nj}}{2 \, {}^a \zeta_j} , \qquad {}^a k_{v_j} = 2 \, {}^a \zeta_j \, {}^a \omega_{nj} \frac{{}^a J_j}{{}^a g_{\tau_j}} \tag{57}$$

The closed-loop performances are chosen with the desired 2 poles of a second order transfer function characterized by, ${}^d \omega_{nj}$, ${}^d \zeta_j$, where:

${}^d \omega_{nj}$ is the desired natural frequency,

${}^d \zeta_j$ is the desired damping coefficient.

Because the actual values are unknown, the gains are calculated from (57), where the unknown actual values, ${}^a \omega_{nj}$, ${}^a \zeta_j$, ${}^a J_j$, ${}^a g_{i_j}$, are replaced respectively by their desired values, ${}^d \omega_{nj}$, ${}^d \zeta_j$, and by their a priori values, ${}^{ap} J_j$, ${}^{ap} g_{\tau_j}$:

 

$$^{a}k_{p_j} = \frac{^{d}\omega_{nj}}{2\,^{d}\zeta_j}\;,\quad ^{a}k_{v_j} = 2\,^{d}\zeta_j\,^{d}\omega_{nj}\frac{^{ap}J_j}{^{ap}g_{\tau_j}} \tag{58}$$

where:

$^{ap}J_j$ and $^{ap}g_{\tau_j}$ are a priori values of the actual unknown values $^{a}J_j$ and $^{a}g_{\tau_j}$, respectively.

Now, let us consider the joint $j$ PD control of the simulated robot which is illustrated Fig. 7.

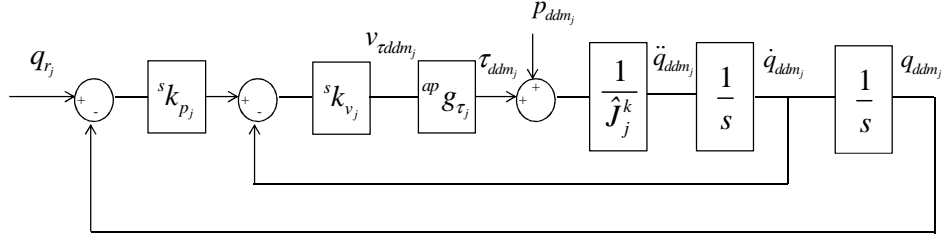

Fig. 7. Joint PD control of the simulated robot.

The variables $\left(v_{\tau ddm_j},\,\tau_{ddm_j},\,q_{ddm_j},\,\dot{q}_{ddm_j},\,\ddot{q}_{ddm_j}\right)$, in Fig. 7, are computed by numerical integration of $DDM(\hat{\chi}^k)$, (21).

The control law of the simulated robot has the same structure as the actual one, Fig. 6, where we take:

$^{a}g_{i_j} = {^{ap}g_{i_j}}$, the a priori value of $^{a}g_{i_j}$,

$^{a}J_j = \hat{J}_j^k$, the value of $J_j$, (51), calculated with the estimation $\hat{\chi}_k$, at iteration k.

$^{s}k_{p_j}$, $^{s}k_{v_j}$ are the gains of the simulated control law.

They are calculated in order to keep the same performances for the simulated closed-loop and for the actual closed-loop, that is to say to keep the same desired values, $^{d}\omega_{nj}$ and $^{d}\zeta_j$, for the closed-loop poles. Then, it becomes:

$$^{s}k_{p_j} = \frac{^{d}\omega_{nj}}{2\,^{d}\zeta_j} = {^{a}k_{p_j}}\;,\quad ^{s}k_{v_j} = 2\,^{d}\zeta_j\,^{d}\omega_{nj}\frac{\hat{J}_j^k}{^{ap}g_{\tau_j}} \tag{59}$$

The proportional gain, $^{s}k_{p_j}$, does not depend at all on the parameters values, but the derivative gain in the simulator, $^{s}k_{v_j}$, must be updated with $\hat{J}_j^k$, at each iteration $k$.

It is important to note that only the gain in the simulated closed-loop, $^{s}k_{v_j}$, is modified during the iterative procedure. The actual gain of the robot control law, $^{a}k_{v_j}$, is not modified.

The simulated closed-loop tuning given by, $^{d}\omega_{nj}$, $^{d}\zeta_j$, differs from the actual one, $^{a}\omega_{nj}$, $^{a}\zeta_j$, with the

 

following ratio, calculated by taking (58) into (56):

$$\frac{{}^a\omega_{nj}}{{}^d\omega_{nj}} = \frac{{}^a\zeta_j}{{}^d\zeta_j} = \sqrt{\frac{{}^{ap}J_j}{{}^aJ_j}\frac{{}^ag_{\tau_j}}{{}^{ap}g_{\tau_j}}} \tag{60}$$

Usually this ratio is between 0.8 and 1.2. The actual values, ${}^a\omega_{nj}$, ${}^a\zeta_j$, can be estimated from step response or frequency analysis of the actual closed-loop. But this is not necessary, because there is little effect on the identification accuracy, assuming, ${}^d\omega_{nj}$, is regularly chosen more than 10 times greater than the frequency range of the robot dynamics.

This allows to keep $(q_{ddm}(\hat{\chi}_k), \dot{q}_{ddm}(\hat{\chi}_k), \ddot{q}_{ddm}(\hat{\chi}_k))$ $(q, \dot{q}, \ddot{q})$, at each iteration k.

We propose to take a regular inertia matrix $M(q_{ddm}, \hat{\chi}^0)$, in order to have a good initialization for the numerical integration of the DDM (21). This is named the "regular initialization".

It can be obtained with:

$$\hat{\chi}^0 = 0, \text{ except for, } Ia_j^0 = 1, j = 1, n \tag{61}$$

The inertia of the rotor and gear of actuator $j$ is generally taken into account in the IDM model (1) as:

$$\tau_{r_j} = Ia_j\, \ddot{q}_j$$

Then, the initial inertia matrix becomes the identity matrix, which is the best regular matrix:

$$M(q_{ddm}, \hat{\chi}^0) = I_n \tag{62}$$

Another simple regular initialization is to take:

$$\hat{\chi}^0 = 0, \text{ except for, } ZZ_j^0 = 1, j = 1, n \tag{63}$$

The initial inertia matrix, $M(q_{ddm}, \hat{\chi}^0)$, is no more the identity matrix, but remains regular.

Another point is to choose the state initial condition of the state vector, $(q_{ddm}(0), \dot{q}_{ddm}(0))$, in order to integrate the DDM (21). Because DIDIM doesn't need the joint position measurement, the actual values $(q(0), \dot{q}(0))$, are supposed to be unknown and we choose, $(q_{ddm}(0), \dot{q}_{ddm}(0)) = (q_r(0), \dot{q}_r(0))$, which is close to $(q(0), \dot{q}(0))$. Because the closed-loop transient response due to different initial conditions differs between the actual and the simulated signals during a transient period of approximately, $5/{}^d\omega_n$, the corresponding joint force/torque samples are eliminated from the identification data in (45).

## V. CASE STUDY: MODELING OF THE SCARA ROBOT

The identification method is carried out on a 2 degree-of-freedom planar direct drive prototype robot without gravity effect, shown in Fig. 8. This direct drive prototype is very suitable for the study of DIDIM because it emphasizes non linear coupling torques while this non-linear effect is divided by at least 2500 for industrial robots with gear ratio greater than 50. Moreover, the dynamic model of this robot depends on eight parameters only, which facilitates the study of the identification efficiency with respect to several conditions. At last, this robot and its real parameters, called the nominal parameters, are well known. Thus, we can check the physical meaning of the identified parameters.

The description of the geometry of the robot uses the modified Denavit and Hartenberg (DHM) notations [30] which are illustrated in Fig.9. The robot is direct driven by 2 DC permanent magnet motors supplied by PWM amplifiers.

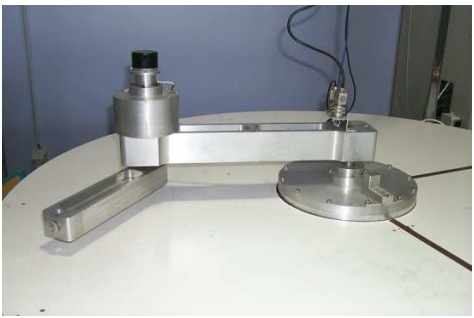
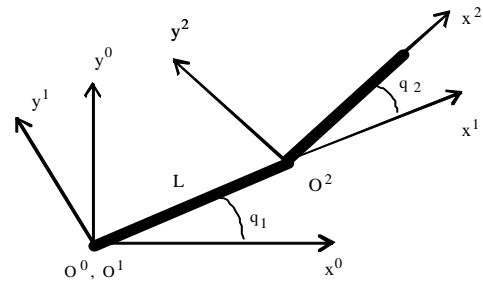

Fig. 8. The scara robot prototype.     Fig. 9. DHM frames of the scara robot.

The dynamic model depends on 8 minimal dynamic parameters, considering 4 friction parameters:

$$\chi = [ZZ_{1R} \quad Fv_1 \quad Fc_1 \quad ZZ_{2R} \quad LMX_2 \quad LMY_2 \quad Fv_2 \quad Fc_2]^T \quad (64)$$

$$ZZ_{1R} = ZZ_1 + Ia_1 + M_2 L^2$$

$$ZZ_{2R} = ZZ_2 + Ia_2$$

$L$=0.5m, is the length of the first link.

In the case of the SCARA robot, the parameters, $LMX_2$, and $LMY_2$, are identified instead of, $MX_2$, and $MY_2$, respectively.

The 8 columns, $IDM_{:,k}$, $k=1,8$, of $IDM(q,\dot{q},\ddot{q})$, in IDIM (7), are the following:

$$IDM_{:,1} = IDM_{ZZ_{1R}} = \begin{bmatrix} \ddot{q}_1 \\ 0 \end{bmatrix}, \quad IDM_{:,2} = IDM_{Fv_1} = \begin{bmatrix} \dot{q}_1 \\ 0 \end{bmatrix},$$

$$IDM_{:,3} = IDM_{Fc_1} = \begin{bmatrix} sign(\dot{q}_1) \\ 0 \end{bmatrix}, \quad IDM_{:,4} = IDM_{ZZ_{2R}} = \begin{bmatrix} \ddot{q}_1 + \ddot{q}_2 \\ \ddot{q}_1 + \ddot{q}_2 \end{bmatrix},$$

$$IDM_{:,5} = IDM_{LMX_2} = \begin{bmatrix} (2\ddot{q}_1 + \ddot{q}_2)\cos q_2 - \dot{q}_2(2\dot{q}_1 + \dot{q}_2)\sin q_2 \\ \ddot{q}_1 \cos q_2 + \dot{q}_1^2 \sin q_2 \end{bmatrix}, \quad (65)$$

$$IDM_{:,6} = IDM_{LMY_2} = \begin{bmatrix} -(2\ddot{q}_1 + \ddot{q}_2)\sin q_2 - \dot{q}_2(2\dot{q}_1 + \dot{q}_2)\cos q_2 \\ \dot{q}_1^2 \cos q_2 - \ddot{q}_1 \sin q_2 \end{bmatrix},$$

$$IDM_{:,7} = IDM_{Fv_2} = \begin{bmatrix} 0 \\ \dot{q}_2 \end{bmatrix}, \quad IDM_{:,8} = IDM_{Fc_2} = \begin{bmatrix} 0 \\ sign(\dot{q}_2) \end{bmatrix}$$

The closed-loop control is a PD control law (53), according to Fig. 6, with:

$J_1 = ZZ_{1R} + ZZ_{2R} + 2 LMX_2$, and $J_2 = ZZ_{2R}$.

The actual gains are calculated with (58), taking a desired damping, $^d\zeta_j = 1$, for joint 1 and joint 2.

The desired natural frequency, $^d\omega_{nj}$, is chosen according to the driving capacity without saturation of the joint drive. For this robot we obtain a full bandwidth with, $^d\omega_{n_1}^f = 1 rd/s$, and $^d\omega_{n_2}^f = 10 rd/s$.

The sample rates of the control and of the measurement are equal to, $f_m = 200$Hz.

Torque data are obtained from (54), and from the current reference data $v_\tau$.

The simulation of the robot is carried out with the same reference trajectory and with the same control law structure as the actual robot.

The gains in the simulator are calculated with (59) and with the same values, $^d\zeta_j = 1$, $^d\omega_{n_1} = 1 rd/s$, and $^d\omega_{n_2} = 10 rd/s$.

## VI. EXPERIMENTAL IDENTIFICATION RESULTS

The new identification process is performed in different cases in order to compare the previous IDIM technique to the new DIDIM technique and to investigate the robustness of DIDIM with respect to the initialization, to the acquisition sampling rate, to the data filtering and to the closed-loop tuning.

All the results are given in SI units, on the joint side.

### A. Comparison of IDIM and DIDIM with good initial values, $\hat{\chi}^0 = \hat{\chi}^{IDIM}$.

At first, the algorithm is initialized with, $\hat{\chi}^{IDIM}$, the vector of parameters identified with the IDIM LS estimator.

The IDIM LS off-line estimation is carried out with a filtered position $\hat{q}$, calculated with a 20Hz cut-off frequency forward and reverse Butterworth filter, and with the velocities $\hat{\dot{q}}$, and the accelerations, $\hat{\ddot{q}}$, calculated with a central difference algorithm of $\hat{q}$. The parallel decimation of $Y_{fm}$ and $W_{fm}$, in (10), is carried out with a sample rate divided by a factor, $n_d=20$, and a lowpass filter cut-off frequency equal to, $0.8*f_m/(2*n_d)=4Hz$.

The results are given in Table 1. It needs only 2 steps to obtain the optimal solution which is very close to the IDIM solution. Hence, the DIDIM method does not improve the IDIM solution calculated with good bandpass filtered data.

A validation is plotted on Fig. 10, at the frequency measurement, $f_m=200Hz$. It shows that the actual joint torques, $Y_{fm}(\tau)$, and the torques estimated with the identified model, $Y_e = W_{\delta fm}(q_{ddm}, \dot{q}_{ddm}, \ddot{q}_{ddm}, \hat{\chi}^2)\hat{\chi}^2$, as defined in (45), (46), (47), are very close.

Both methods give a small relative norm error, $\|Y - W\hat{\chi}\| / \|Y\| < 3\%$, which shows a good accuracy for the model and for the identified values.

It can be seen that the parameters, $Fv_1$, and $Fv_2$, have no significant estimations because of their large relative standard deviation (>30%). They have no significant contribution in the joint torques and they can be cancelled to keep a set of essential parameters of a simplified dynamic model, without loss of accuracy [31].

However, we prefer to keep all the parameters in the following, for a better comparison of IDIM and DIDIM identification methods.

TABLE 1: COMPARISON OF IDIM AND DIDIM METHODS

| Parameter | IDIM | | | DIDIM | | | |
|---|---|---|---|---|---|---|---|
| | $\hat{\chi}^{IDIM}$ | $2\sigma_{\hat{\chi}}$ | $\%\sigma_{\hat{\chi}_r}$ | $\hat{\chi}^0 = \hat{\chi}^{IDIM}$ | $\hat{\chi}^2$ | $2\sigma_{\hat{\chi}}$ | $\%\sigma_{\hat{\chi}_r}$ |
| $ZZ_{1R}$ | 3.44 | 0.034 | 0.50 | 3.44 | 3.45 | 0.036 | 0.52 |
| $Fv_1$ | 0.03 | 0.031 | 52.0 | 0.03 | 0.04 | 0.032 | 40.0 |
| $Fc_1$ | 0.82 | 0.1 | 6.0 | 0.82 | 0.82 | 0.05 | 3.0 |
| $ZZ_2$ | 0.062 | 0.0006 | 0.51 | 0.062 | 0.061 | 0.0006 | 0.49 |
| $LMX_2$ | 0.121 | 0.0014 | 0.56 | 0.121 | 0.124 | 0.0013 | 0.52 |
| $LMY_2$ | 0.007 | 0.0007 | 5.0 | 0.007 | 0.007 | 0.0005 | 3.5 |
| $Fv_2$ | 0.013 | 0.006 | 23.0 | 0.013 | 0.014 | 0.0084 | 30.0 |
| $Fc_2$ | 0.137 | 0.006 | 2.30 | 0.137 | 0.133 | 0.0080 | 3.0 |
| $\|Y - W\hat{\chi}^{IDIM}\| / \|Y\|$=0.024 | | | | $\|Y - W\hat{\chi}^2\| / \|Y\|$=0.021 | | | |

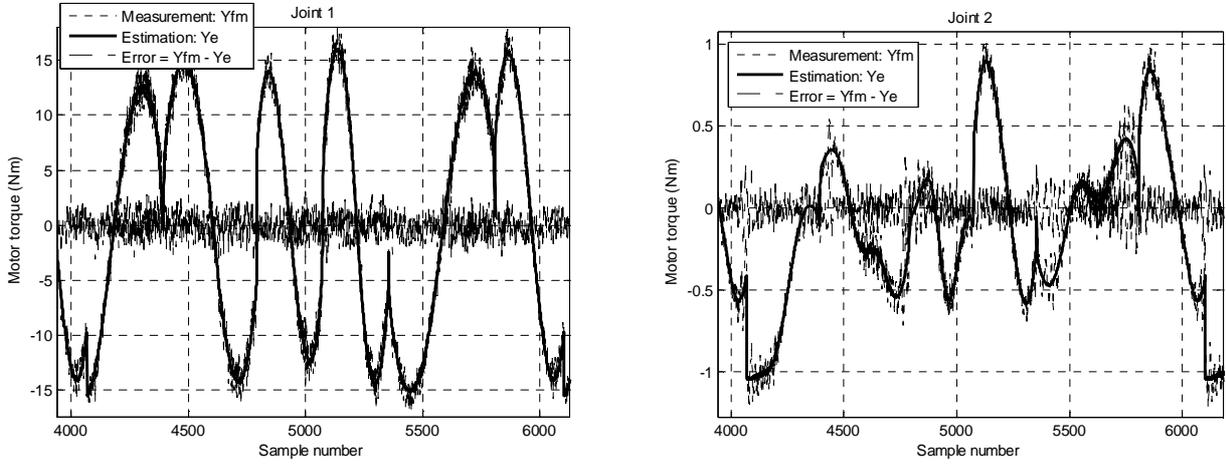

Fig. 10. DIDIM, validation, $Y_e = W_{\delta fm}(q_{ddm}, \dot{q}_{ddm}, \ddot{q}_{ddm}, \hat{\chi}^2)\hat{\chi}^2$.

*B. DIDIM, validation of the regular initialization, $M(q_{ddm}, \hat{\chi}^0) = I_2$*

The robustness of DIDIM with respect to a wrong initialization, such as the regular initialization (62), is investigated.

The initial values of the dynamic parameters are given by (61), with:

$\hat{\chi}^0 = \begin{bmatrix} 1 & 0 & 0 & 1 & 0 & 0 & 0 & 0 \end{bmatrix}^T$

The identified values given in Table 2, are very close to those given in Table 1. This result validates the regular initialization procedure, described in section IV.B.

Moreover the algorithm converges in only 3 steps and is not time consuming.

TABLE 2: DIDIM WITH THE REGULAR INITIALIZATION

| Parameter | $\hat{\chi}^0$ | $\hat{\chi}^3$ | $2\sigma_{\hat{\chi}}$ | $\%\sigma_{\hat{\chi}_r}$ |
|---|---|---|---|---|
| $ZZ_{1R}$ | 1 | 3.45 | 0.014 | 0.2 |
| $Fv_1$ | 0 | 0.02 | 0.012 | 15 |
| $Fc_1$ | 0 | 0.85 | 0.016 | 1.0 |
| $ZZ_2$ | 1 | 0.061 | 0.0001 | 0.1 |
| $LMX_2$ | 0 | 0.124 | 0.0002 | 0.1 |
| $LMY_2$ | 0 | 0.007 | 0.0003 | 2.0 |
| $Fv_2$ | 0 | 0.01 | 0.003 | 10 |
| $Fc_2$ | 0 | 0.132 | 0.0008 | 0.3 |

The relative norm errors on joint position, velocity and acceleration are plotted in Fig.11 ,with the following legend:

• norm error relative to the actual filtered joint position, $\|q_{ddm_j} - \hat{q}_j\|/\|\hat{q}_j\|$, velocity $\|\dot{q}_{ddm_j} - \hat{\dot{q}}_j\|/\|\hat{\dot{q}}_j\|$, and acceleration, $\|\ddot{q}_{ddm_j} - \hat{\ddot{q}}_j\|/\|\hat{\ddot{q}}_j\|$, where $(\hat{q}, \hat{\dot{q}}, \hat{\ddot{q}})$, are calculated as given in section VI.A.

* norm error relative to the reference joint position, $\|q_{ddm_j} - q_{r_j}\|/\|q_{r_j}\|$, velocity, $\|\dot{q}_{ddm_j} - \dot{q}_{r_j}\|/\|\dot{q}_{r_j}\|$, and acceleration, $\|\ddot{q}_{ddm_j} - \ddot{q}_{r_j}\|/\|\ddot{q}_{r_j}\|$.

The assumption (40), made in section IV.B, $(q_{ddm}(\hat{\chi}_k), \dot{q}_{ddm}(\hat{\chi}_k), \ddot{q}_{ddm}(\hat{\chi}_k))$ $(\hat{q}, \hat{\dot{q}}, \hat{\ddot{q}})$, at each iteration k, is confirmed on Fig.11. , with a constant relative norm error close to 0.5% for the position, 5%, for the velocity and 10%, for the acceleration.

These results validate the updating procedure (59), of the simulated PD control law gains.

It can be seen also on Fig.11. , that the simulated trajectory, $(q_{ddm}(\hat{\chi}_k), \dot{q}_{ddm}(\hat{\chi}_k), \ddot{q}_{ddm}(\hat{\chi}_k))$, is 3 to 5 times closer to the actual one, $(\hat{q}, \hat{\dot{q}}, \hat{\ddot{q}})$, than to the reference one, $(q_r, \dot{q}_r, \ddot{q}_r)$, with a relative norm error close to 1.5% for the position, 15%, for the velocity and 30%, for the acceleration. Moreover, this error depends on the closed-loop bandwidth. It means that computing the observation matrix in (13) with the reference trajectory, $(q_r, \dot{q}_r, \ddot{q}_r)$, leads to a bad identification of the dynamic parameters.

Then, the right assumption is, $(q_{ddm}(\hat{\chi}_k), \dot{q}_{ddm}(\hat{\chi}_k), \ddot{q}_{ddm}(\hat{\chi}_k))$ $(\hat{q}, \hat{\dot{q}}, \hat{\ddot{q}})$, (40), made in section IV.B.

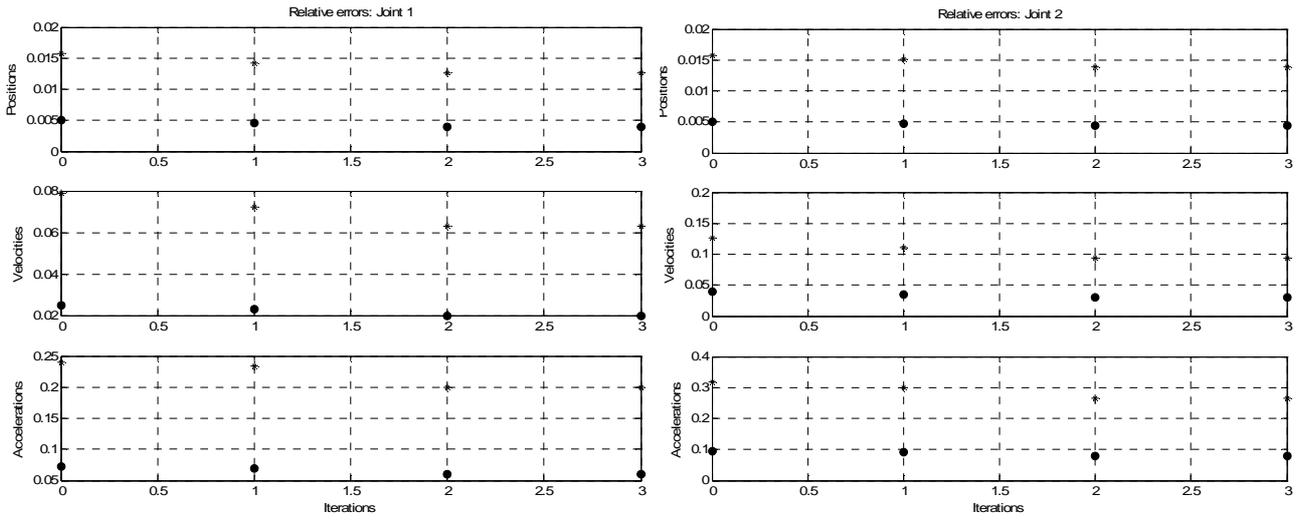

Fig.11. • norm error relative to the filtered actual position, velocity, acceleration.

\* norm error relative to the reference position, velocity, acceleration.

We have seen that $(q_{ddm}(\hat{\chi}_k), \dot{q}_{ddm}(\hat{\chi}_k), \ddot{q}_{ddm}(\hat{\chi}_k))$ $(\hat{q}, \hat{\dot{q}}, \hat{\ddot{q}})$, at each iteration k, with a constant small error. On the contrary, the relative torque norm error, given in Table 3, and plotted on Fig. 12, dramatically decreases in only 3 steps. This shows the fast algorithm convergence.

TABLE 3: RELATIVE NORM ERROR OF JOINT TORQUE: $\|Y^j - W^j \hat{\chi}^k\| / \|Y^j\|$

| Iteration $k$ | 0 | 1 | 2 | 3 |
|---|---|---|---|---|
| Joint $j=1$ | 0.42 | 0.036 | 0.02 | 0.018 |
| Joint $j=2$ | 3.20 | 0.110 | 0.02 | 0.022 |

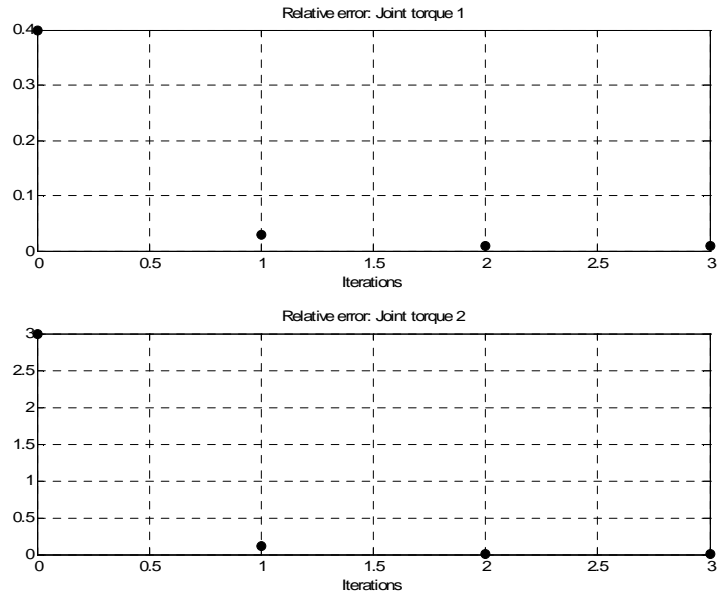

Fig. 12. DIDIM, convergence of the joint torque error, $\|Y^j - W^j \hat{\chi}^k\| / \|Y^j\|$

The fast convergence of each parameter is shown in Table 4 , and is plotted on Fig. 13.

TABLE 4: PARAMETERS CONVERGENCE

| Parameters | $\hat{\chi}^0$ | $\hat{\chi}^1$ | $\hat{\chi}^2$ | $\hat{\chi}^3$ |
|---|---|---|---|---|
| $ZZ_{1R}$ | 1 | 3.46 | 3.45 | 3.45 |
| $Fv_1$ | 0 | 0.04 | 0.02 | 0.02 |
| $Fc_1$ | 0 | 0.86 | 0.85 | 0.85 |
| $ZZ_2$ | 1 | 0.06 | 0.061 | 0.061 |
| $LMX_2$ | 0 | 0.122 | 0.124 | 0.124 |
| $LMY_2$ | 0 | 0.05 | 0.07 | 0.07 |
| $Fv_2$ | 0 | 0.005 | 0.01 | 0.01 |
| $Fc_2$ | 0 | 0.130 | 0.132 | 0.132 |

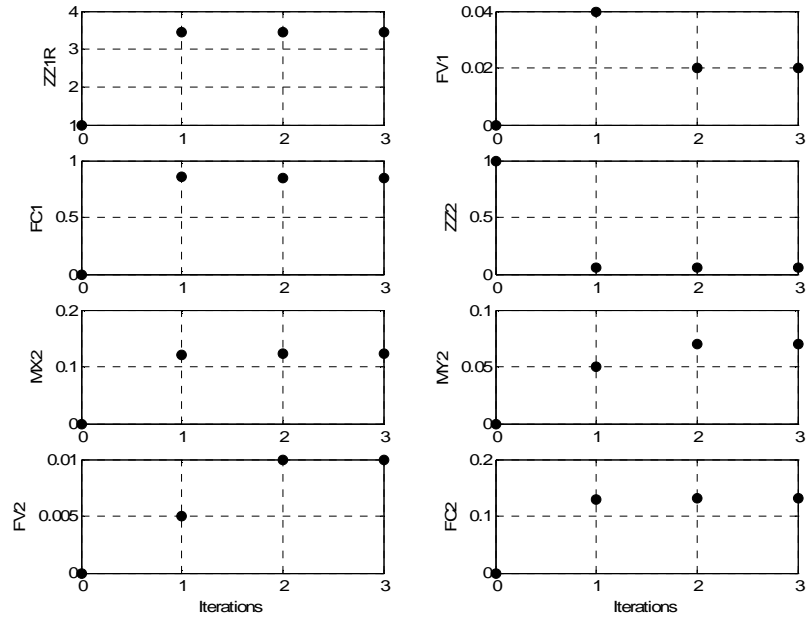

Fig. 13. DIDIM, parameters convergence

*C. Comparison of IDIM and DIDIM robustness with respect to a low sample rate.*

The actual torque and the simulated data are resampled to obtain a low frequency measurement $f_m = 0.5$Hz. This is a downsample procedure without lowpass antialiasing filtering which investigates a real problem on industrial robots where the available sample rate measurement given by the controller may be much lower than the control sample rate. All the actual and simulated data are sampled at $f_m = 0.5$Hz.

The IDIM LS estimation is carried out with the measured joint position $q$, and with $(\hat{\dot{q}}, \hat{\ddot{q}})$, calculated by a central difference algorithm of $q$, without lowpass Butterworth filtering. There is no parallel

 

decimation. DIDIM starts with the regular initialization. Results are given in Table 5.

TABLE 5: IDIM AND DIDIM, LOW SAMPLING RATE, $f_m = 0.5$Hz.

| Parameter | IDIM | | | DIDIM | | | |
|---|---|---|---|---|---|---|---|
| | $\hat{\chi}^{IDIM}$ | $2\sigma_{\hat{\chi}}$ | %$\sigma_{\hat{\chi}_r}$ | $\hat{\chi}^0$ | $\hat{\chi}^3$ | $2\sigma_{\hat{\chi}}$ | %$\sigma_{\hat{\chi}_r}$ |
| $ZZ_{1R}$ | 3.10 | 0.03 | 0.3 | 1.0 | 3.45 | 0.04 | 0.5 |
| $Fv_1$ | 0.9 | 1.8 | 100 | 0 | 0.04 | 0.02 | 30 |
| $Fc_1$ | 1.0 | 0.1 | 5 | 0 | 0.81 | 0.05 | 3 |
| $ZZ_2$ | 0.025 | 0.003 | 5.5 | 1.0 | 0.061 | 0.0006 | 0.5 |
| $LMX_2$ | 0.075 | 0.008 | 5.3 | 0 | 0.124 | 0.001 | 0.5 |
| $LMY_2$ | -0.02 | 0.01 | 250 | 0 | 0.008 | 0.0006 | 4.0 |
| $Fv_2$ | 0.35 | 5.6 | 800 | 0 | 0.01 | 0.005 | 25 |
| $Fc_2$ | 0.19 | 0.087 | 23 | 0 | 0.13 | 0.008 | 3.0 |
| $\|Y - W\hat{\chi}^{IDIM}\| / \|Y\| = 0.5$ | | | | $\|Y - W\hat{\chi}^3\| / \|Y\| = 0.04$ | | | |

The identified values with IDIM are not good while the identified values with DIDIM are still good. This shows the robustness of DIDIM with respect to the sampling rate measurement.

IDIM fails because there is an amplitude distortion in the estimation of $(\hat{\dot{q}}, \hat{\ddot{q}})$, with a central difference of $q$, sampled at a too low frequency $f_m$. This point is illustrated in Table 6, which gives the relative norm errors on velocity (80%) and acceleration (80%).

$(\hat{\dot{q}}(200\text{Hz}), \hat{\ddot{q}}(200\text{Hz}))$, is calculated from $q$, sampled at 200Hz and lowpass filtered at a 0.5Hz cut-off frequency, and derived with a central difference algorithm.

$(\hat{\dot{q}}(0.5\text{Hz}), \hat{\ddot{q}}(0.5\text{Hz}))$, is calculated from $q$, sampled at 0.5Hz and derived with a central difference algorithm.

DIDIM succeeds because, $(q_{ddm}, \dot{q}_{ddm}, \ddot{q}_{ddm})$, is computed with accuracy by the integration of the DDM with a well-tuned variable step solver, and it can be sampled without error at any frequency $f_m$.

TABLE 6: IDIM, JOINT DATA ERRORS AT $f_m = 0.5$Hz

| | |
|---|---|
| $\|\hat{\dot{q}}_1(200\text{Hz}) - \hat{\dot{q}}_1(0.5\text{Hz})\| / \|\hat{\dot{q}}_1(200\text{Hz})\|$ | 0.39 |
| $\|\hat{\ddot{q}}_1(200\text{Hz}) - \hat{\ddot{q}}_1(0.5\text{Hz})\| / \|\hat{\ddot{q}}_1(200\text{Hz})\|$ | 0.73 |
| $\|\hat{\dot{q}}_2(200\text{Hz}) - \hat{\dot{q}}_2(0.5\text{Hz})\| / \|\hat{\dot{q}}_2(200\text{Hz})\|$ | 0.80 |
| $\|\hat{\ddot{q}}_2(200\text{Hz}) - \hat{\ddot{q}}_2(0.5\text{Hz})\| / \|\hat{\ddot{q}}_2(200\text{Hz})\|$ | 0.81 |

 

*D. Comparison of IDIM and DIDIM, without data filtering.*

All the actual and simulated data are sampled at $f_m$ = 200Hz.

The IDIM LS estimation is carried out with the measured joint position $q$, and with $(\hat{\dot{q}},\hat{\ddot{q}})$, calculated by a central difference algorithm of $q$, without lowpass Butterworth filtering. There is no parallel decimation. DIDIM starts with the regular initialization. Results are given in Table 7.

TABLE 7: IDIM AND DIDIM ESTIMATION WITHOUT DATA FILTERING

| Parameter | IDIM | | | DIDIM | | | |
|---|---|---|---|---|---|---|---|
| | $\hat{\chi}^{IDIM}$ | $2\sigma_{\hat{\chi}}$ | $\%\sigma_{\hat{\chi}_r}$ | $\hat{\chi}^0$ | $\hat{\chi}^2$ | $2\sigma_{\hat{\chi}}$ | $\%\sigma_{\hat{\chi}_r}$ |
| $ZZ_{1R}$ | 1.50 | 0.05 | 1.60 | 1.0 | 3.45 | 0.007 | 0.1 |
| $Fv_1$ | 0.095 | 0.15 | 80.0 | 0 | 0.05 | 0.023 | 21 |
| $Fc_1$ | 0.55 | 0.26 | 23.3 | 0 | 0.81 | 0.004 | 0.24 |
| $ZZ_2$ | 0.14 | 0.018 | 6.7 | 1.0 | 0.061 | 0.0004 | 0.3 |
| $LMX_2$ | 0.63 | 0.035 | 2.7 | 0 | 0.124 | 0.0015 | 0.3 |
| $LMY_2$ | 0.1 | 0.023 | 11.8 | 0 | 0.008 | 0.0009 | 5.6 |
| $Fv_2$ | 0.001 | 0.143 | 700.0 | 0 | 0.023 | 0.0022 | 48 |
| $Fc_2$ | 0.19 | 0.244 | 68.40 | 0 | 0.13 | 0.0038 | 1.5 |
| | $\|Y-W\hat{\chi}^{IDIM}\|/\|Y\|$=0.8 | | | $\|Y-W\hat{\chi}^2\|/\|Y\|$=0.08 | | | |

The identified values with IDIM are not good while the identified values with DIDIM are still good.

IDIM fails because of the too large noise in the observation matrix, $W_{fm}(q,\hat{\dot{q}},\hat{\ddot{q}})$, coming from the derivation of $q$, without lowpass filtering. Then the LS estimation is biased.

DIDIM succeeds because the observation matrix, $W_{\delta fm}(q_{ddm},\dot{q}_{ddm},\ddot{q}_{ddm},\hat{\chi}^k)$, is calculated without noise with the simulated values $(q_{ddm},\dot{q}_{ddm},\ddot{q}_{ddm})$.

This validation shows that DIDIM cancels the bias of IDIM estimation, coming from a noisy estimation of $(\hat{q},\hat{\dot{q}},\hat{\ddot{q}})$, which gives a too noisy observation matrix $W_{fm}(q,\hat{\dot{q}},\hat{\ddot{q}})$.

*E. DIDIM robustness with respect to error in the simulated closed-loop tuning, ${}^d\omega_n$*

This section investigates the effect of an error between the actual value, ${}^a\omega_n$, and the simulated value ${}^d\omega_n$, of the natural frequency which represents the closed-loop bandwidth.

The DIDIM identification is performed taking half the values of the full ones given in section V, ${}^d\omega_{n_1} = {}^d\omega_{n_1}^f/2$=1/2 (rd/s) and ${}^d\omega_{n_2} = {}^d\omega_{n_2}^f/2 = 10/2$ (rd/s), and the same procedure used to obtain results shown in Table 2, that is to say a frequency measurement, $f_m$=200Hz, and a parallel decimation with a factor, $n_d$=20, and a lowpass filter cut-off frequency equal to 4Hz.

The parameters, given in Table 8, converge in only 6 steps to values which are very close to those obtained in Table 2, with a full closed-loop bandwidth.

TABLE 8: DIDIM, WITH SIMULATED HALF FULL BANDWIDTH, $^d\omega_n = {^d\omega_n^f}/2$

| Parameter | $\hat{\chi}^0$ | $\hat{\chi}^6$ | $2\sigma_{\hat{\chi}}$ | $\%\sigma_{\hat{\chi}_r}$ |
|---|---|---|---|---|
| $ZZ_{1R}$ | 1 | 3.44 | 0.014 | 0.2 |
| $Fv_1$ | 0 | 0.02 | 0.012 | 15 |
| $Fc_1$ | 0 | 0.86 | 0.016 | 1.0 |
| $ZZ_2$ | 1 | 0.060 | 0.0001 | 0.1 |
| $LMX_2$ | 0 | 0.124 | 0.0002 | 0.1 |
| $LMY_2$ | 0 | 0.007 | 0.0003 | 2.0 |
| $Fv_2$ | 0 | 0.01 | 0.003 | 10 |
| $Fc_2$ | 0 | 0.13 | 0.0008 | 0.3 |

The relative norm errors on joint position, velocity and acceleration are plotted on Fig. 14, with the same legend as those given for Fig.11, section VI.B.

It can be seen that, $(q_{ddm}(\hat{\chi}_k), \dot{q}_{ddm}(\hat{\chi}_k), \ddot{q}_{ddm}(\hat{\chi}_k))$ $(\hat{q}, \dot{\hat{q}}, \ddot{\hat{q}})$, at each iteration k, with a constant norm error larger but close to the value obtained with the full bandwidth, Fig.11, close to, 0.5% for the position, 5%, for the velocity and 10%, for the acceleration.

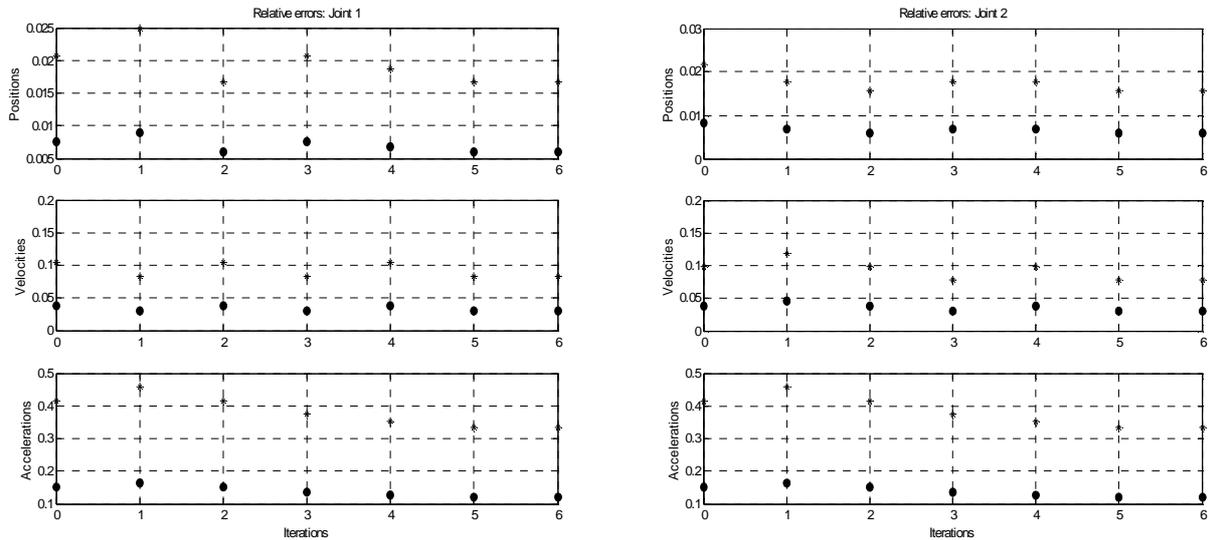

Fig. 14. ● norm error relative to the filtered actual position, velocity, acceleration.

\* norm error relative to the reference position, velocity, acceleration.

 

The relative torque norm error, given in Table 9, and plotted in Fig. 15, decreases in 6 steps, only twice more than with the full bandwidth, Table 3, Fig. 12. This shows that DIDIM is not very sensitive to error in the simulated closed-loop bandwidth, provided the control law structure is known.

However, DIDIM fails beyond 1/3 of the full bandwidth, with ${}^d\omega_n \leq {}^d\omega_n^f / 3$.

TABLE 9: RELATIVE NORM ERROR OF JOINT TORQUE, $\left\|Y^j - W^j\hat{\chi}^k\right\| / \left\|Y^j\right\|$, FULL BANDWIDTH/2

| Iteration $k$ | 0   | 1    | 2    | 3    | 4     | 5    | 6    |
|---------------|-----|------|------|------|-------|------|------|
| Joint $j$=1   | 0.6 | 0.05 | 0.06 | 0.04 | 0.025 | 0.02 | 0.02 |
| Joint $j$=2   | 3.0 | 0.11 | 0.05 | 0.02 | 0.025 | 0.02 | 0.02 |

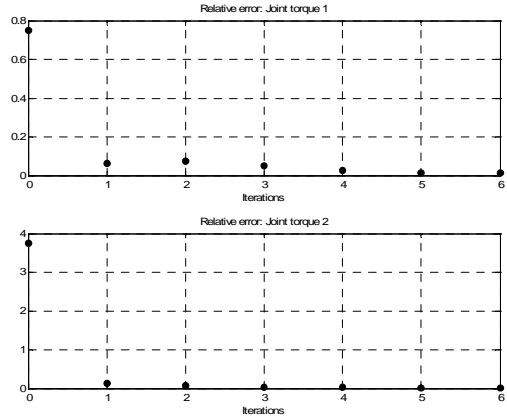

Fig. 15. DIDIM, convergence of the joint torque error, $\left\|Y^j - W^j\hat{\chi}^k\right\| / \left\|Y^j\right\|$

## VII. CONCLUSION

This paper deals with a new off-line identification technique of robot dynamic parameters, called DIDIM for Direct and Inverse Dynamic Identification Models technique. This method is a closed-loop Output Error approach, but considering the output is no more the joint position but the joint force/torque. The optimal parameters are the solution of a non-linear least-squares problem which is solved with a Gauss-Newton method. Each step of the iterative procedure of the Gauss-Newton regression is dramatically simplified to a linear regression which is solved with the Inverse Dynamic Identification Model technique (IDIM). Then, DIDIM mixes the closed-loop OE technique and the IDIM technique.

DIDIM needs a closed-loop simulation of the robot using the direct dynamic model (DDM) and assuming the same structure of the control law and the same reference trajectory for both the actual and the simulated robot. Then, it needs to initialize the parameters and the state vector of the DDM.

The difficulties for the choice of the initial conditions for non-linear LS problem are overcome with a "regular initialization" of the parameters and an updating of the control law gains at each step of the iterative procedure. The initial state is given by the initial values of the reference trajectory.

An experimental validation is carried out on a 2 dof robot. The following points were checked:

- DIDIM gives the same results as IDIM, provided well-tuned data filtering for IDIM, adapted to the system dynamics,
- DIDIM is robust to the initialization of both parameters and state,
- DIDIM is robust to the closed-loop performances tuning errors between the simulated and the actual closed-loop robot, provided the same control law structure.

Compared to IDIM, DIDIM technique is particularly attractive thanks to the following reasons:

- It needs only the actuator force/torque measurement or estimation,
- It avoids tuning the bandpass filter in the IDIM method by using the integration of the DDM in a closed-loop simulation where the tuning of the bandwidth automatically defines the same frequency range for the dynamics of the actual system and of the model to be identified,
- It cancels bias in IDIM due to errors in bandpass filtering data, or no filtering at all, or too low frequency measurement,
- It combines the inverse and the direct dynamic model and validates, in the same identification procedure, both models for computed torque control and for simulation. Up to now, the DDM was validated *a posteriori* in simulation.

Future work concerns the validation of DIDIM on a 6 dof industrial robot.

**Maxime Gautier** received the "Doctorat d'Etat" degree in robotics and control engineering from the University of Nantes, Nantes, France, in 1990. Since 1991, he has been a Professor in automatic control with the Université de Nantes. He is carrying out his research with the Robotics Team, "Institut de Recherche en Communications et Cybernétique de Nantes" (IRCCyN). His research topics include modeling, identification, and control of robots.

**Alexandre Janot** was born in France on July 22, 1980. He received the B.Sc. and the M. Sc. in electrical and automation engineering from the University of Nantes and Ecole Centrale de Nantes, France, in 2002 and 2004, respectively. He received the Ph. D. degree from both the University of Nantes and the French Atomic Energy Commission (CEA) in 2007. He joined HAPTION S.A. in 2008, and is now working on the identification and the control of haptic devices.

**Pierre-Olivier Vandanjon** was born in France on November the 4th, 1969. He received the B. Sc and the M. Sc in applied mathematics from the University Paris-Dauphine. He graduated, as a civil engineer, from Ecole des Mines de Paris. He received the Ph. D degree from both the Ecole des Mines de Paris and the French Atomic Energy Commission (CEA) in 1995. He joined the Laboratoire Central des Ponts et Chaussées (LCPC) in 1997, as a junior researcher. He is providing robotics techniques on projects concerning road construction and maintenance.